\documentclass{ceurart}
\conference{The Conference and Labs of the Evaluation Forum (CLEF) 2026}

\usepackage[utf8]{inputenc}
\usepackage[T1]{fontenc}
\usepackage{nicefrac}
\usepackage{microtype}
\usepackage{multirow}
\usepackage{booktabs}
\usepackage{listings}

\lstdefinestyle{promptstyle}{
    basicstyle=\ttfamily\scriptsize,
    breaklines=true,
    breakatwhitespace=true,
    columns=fullflexible,
    frame=single,
    keepspaces=true,
    showstringspaces=false,
    captionpos=b
}

\begin{document}

\title{DistilledGemma: Balanced Efficiency-Accuracy for Person-Place Relation Extraction from Multilingual Historical Articles}

\author[1]{Youssef Aboelwafa}[email=es-YoussefHossamElDin2025@alexu.edu.eg]
\fnmark[1]

\author[1]{Ahmed Samir}[email=es-AhmedAbdelMaksoud2025@alexu.edu.eg]
\fnmark[1]

\author[1]{Nagwa Elmakky}[email=nagwamakky@alexu.edu.eg]

\author[1]{Marwan Torki}[email=mtorki@alexu.edu.eg]

\address[1]{Alexandria University, Egypt}

\fntext[1]{Equal contribution.}

\begin{abstract}
We present \textbf{DistilledGemma}, an efficient and accurate system for the \textbf{HIPE-2026} shared task on person-place relation extraction from multilingual historical newspaper articles in English, German, and French. Our approach adopts a three-stage knowledge distillation pipeline designed to balance classification accuracy with computational efficiency. In the first stage, we systematically explored prompt engineering strategies across eight large language models to identify the most effective reasoning architecture for this challenging task. In the second stage, we applied supervised fine-tuning (SFT) via QLoRA to a \textbf{Gemma~4 26B A4B} teacher model, leveraging its strong multilingual capabilities to generate silver-standard chain-of-thought traces across the training corpus. In the final stage, we performed response-level distillation to transfer these learned reasoning patterns into a compact \textbf{Gemma~4 E2B} student model. To address the class imbalance between positive and negative relation pairs and to enforce semantic consistency, we incorporated rule-based post-processing on all generated outputs. Specifically, this step enforced entity-type compatibility, relation-direction constraints, contextual trigger patterns, and the logical constraint. In the official evaluation, our team \textbf{WHEREAMI} ranked \textbf{3rd} on the standard test set with an accuracy profile mean score of \textbf{0.688}, and \textbf{2nd} on the binary test set with a mean score of \textbf{0.8156}. Notably, by distilling knowledge from the 26B teacher to the 2.3B student, we preserved strong reasoning capabilities while reducing the deployed model size to approximately \textbf{2.3B effective parameters}; the LoRA adapters used during training were merged into the student for inference. This configuration ranked \textbf{2nd} in the balanced efficiency-accuracy profile across both the standard and binary test sets. These results demonstrate that knowledge distillation provides a practical and scalable solution for historical document processing, achieving competitive performance without excessive computational cost.

\end{abstract}

\maketitle

\section{Introduction}
\label{sec:intro}

Historical newspapers constitute invaluable resources for studying past societies, migration patterns, and geopolitical events. A critical step in unlocking this information is \textbf{person-place relation extraction}: determining whether a person mentioned in an article bears a geographic connection to a place also mentioned therein.

\begin{figure}[t]
\centering
\includegraphics[width=0.7\linewidth]{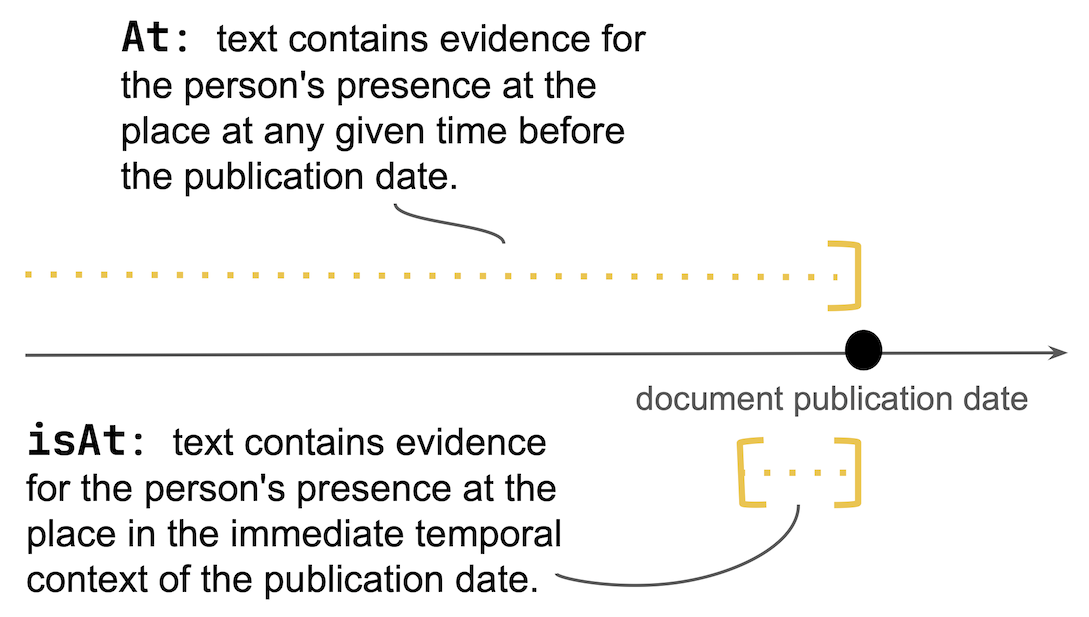}
\caption{Illustration of the distinction between \textit{at} and \textit{isAt} relations.}
\label{fig:at_isat}
\end{figure}

\begin{figure}[t]
\centering
\includegraphics[width=1\linewidth]{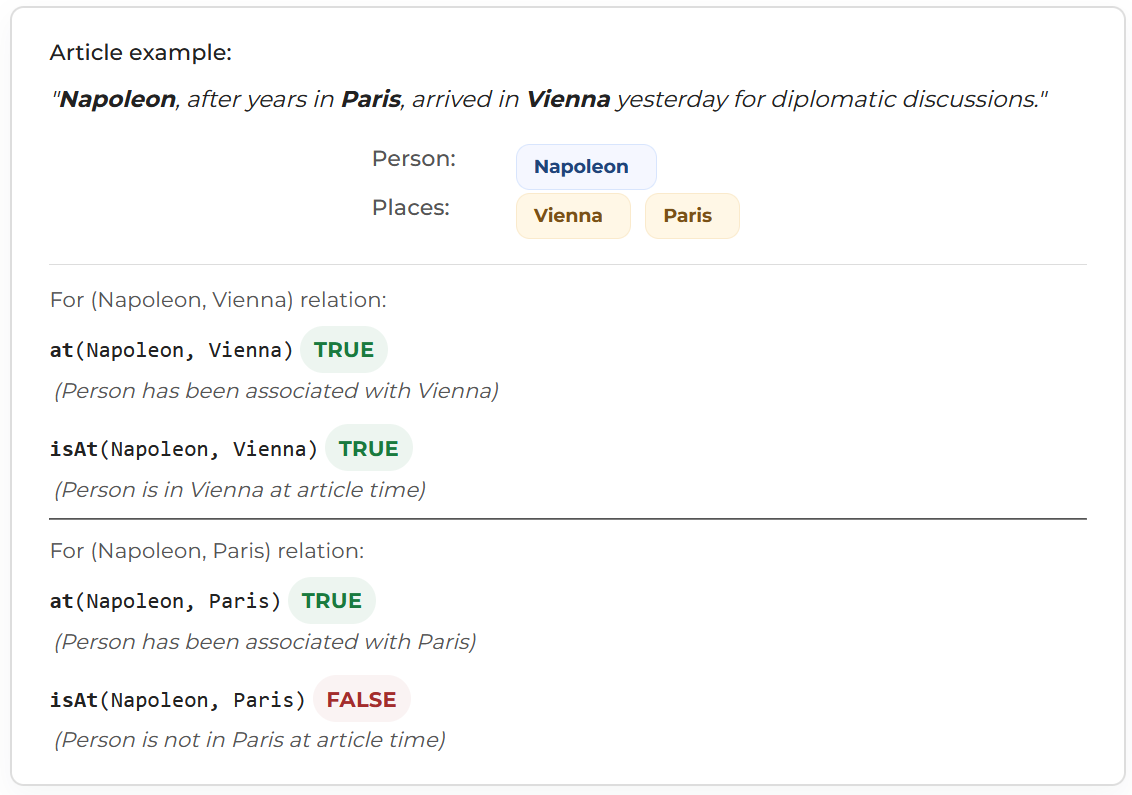}
\caption{Illustration of the distinction between \textit{at} and \textit{isAt} relations in HIPE-2026. The \textit{at} relation captures historical or general associations, while \textit{isAt} denotes the person's location at the time described in the article.}
\label{fig:example}
\end{figure}

The HIPE-2026 shared task~\cite{opitz2026clef, opitz_overview_2026, opitz_extended_2026} formalizes this problem as a classification task over (person, place) pairs extracted from multilingual newspaper articles in English, German, and French. Each pair must be labelled for two relations: \textbf{at} (whether the person has \emph{ever} been associated with the place; \textsc{TRUE}/\textsc{PROBABLE}/\textsc{FALSE}) and \textbf{isAt} (whether the person is at the place \emph{within the immediate temporal horizon of the article}; \textsc{TRUE}/\textsc{FALSE}) as shown in Figure~\ref{fig:at_isat} and~\ref{fig:example}.

This task presents several challenges:
\begin{itemize}
    \item \textbf{Noisy OCR text}: Historical documents contain frequent optical character recognition errors that degrade entity mention quality.
    \item \textbf{Class imbalance}: The \textsc{FALSE} label dominates both relations (${\sim}$55\% for \textit{at}, ${\sim}$78\% for \textit{isAt}), biasing models toward negative predictions.
    \item \textbf{Multilingual reasoning}: Models must perform consistently across three typologically diverse languages with varying training data availability.
    \item \textbf{Implicit evidence}: Many person-place connections are implied rather than explicitly stated, requiring world knowledge and contextual inference.
\end{itemize}

Large language models (LLMs) have demonstrated impressive zero-shot and few-shot capabilities for relation extraction~\cite{wadhwa2023revisiting, wan2023gpt}, but models with sufficient capacity (e.g., 26B+ parameters) are impractical for deployment in digital humanities pipelines that must process millions of documents.
This motivates our central research question: \emph{Can we distill the relation extraction capabilities of a large, fine-tuned LLM into a model small enough for practical deployment without catastrophic performance loss?}

Our contributions are as follows:
\begin{enumerate}
    \item A \textbf{comprehensive evaluation} of eight LLM configurations spanning prompt engineering, zero-shot inference, and supervised fine-tuning for person-place relation extraction across three languages.
    \item A \textbf{three-stage distillation pipeline} that fine-tunes a Gemma~4 26B A4B teacher via QLoRA, generates silver-standard training data with chain-of-thought (CoT) reasoning, and trains a Gemma~4 E2B student on the teacher's outputs.
    \item \textbf{Empirical evidence} that knowledge distillation recovers ${\sim}$88\% of teacher performance, establishing a practical efficiency-accuracy trade-off for historical NLP applications.
\end{enumerate}

\section{Related Work}
\label{sec:related}

\paragraph{Encoder-based relation extraction.}
Before the recent shift to generative LLMs, relation extraction was commonly framed as discriminative classification over marked entity pairs, using either hand-engineered features~\cite{zhang-etal-2017-position} or contextual encoders~\cite{pathak2016context}. These lightweight baselines remain important for low-resource and imbalanced settings because they condition directly on entity spans and require only a small task head. Entity-aware representation learning, including task-agnostic relation representations from entity-linked text~\cite{baldini2019matching}, typed entity-marker models~\cite{zhou2022improved}, and PL-Marker span-pair packing~\cite{ye2022packed}, provides strong baselines for sentence-level RE. Cross-lingual encoders such as XLM-R~\cite{conneau2020xlmr} and entity-aware multilingual encoders such as mLUKE~\cite{ri2022mluke} are especially relevant for multilingual HIPE-style data. For longer article-level evidence, document-level RE benchmarks such as DocRED highlight the need to synthesize evidence across sentences and entity mentions~\cite{yao2019docred}. Recent multilingual and digital-humanities RE work has also used guided distant supervision to build German biographical relation data and evaluate multilingual transfer~\cite{plum2024guided}. Low-resource RE benchmarks show that class balancing, data augmentation, and self-training are useful alternatives, but their gains can be inconsistent under long-tailed label distributions~\cite{xu2022towards}. These approaches are therefore strong lightweight baselines and complementary alternatives to our LLM distillation pipeline.

\paragraph{Relation extraction with LLMs.}
Recent work has demonstrated that LLMs can perform competitive relation extraction through in-context learning~\cite{wan2023gpt} and chain-of-thought prompting~\cite{wei2022chain}, particularly when explicit reasoning steps are elicited before label prediction. Wadhwa et al.~\cite{wadhwa2023revisiting} showed that GPT-class models match or exceed supervised baselines on standard benchmarks, albeit at significantly higher computational cost.

\paragraph{Historical NLP.}
The HIPE shared task series~\cite{hipe2022, ehrmann2023extended} has driven substantial progress in named entity recognition and linking for historical texts, establishing standardized benchmarks across multiple languages and time periods. These tasks highlight unique challenges including OCR noise, archaic language usage, and domain-specific entity types that are absent from modern NLP benchmarks. The HIPE-2026 edition~\cite{opitz_overview_2026} extends the series to person-place relation extraction, introducing new evaluation profiles that jointly consider accuracy and computational efficiency.

\paragraph{Knowledge distillation and teacher alignment.}
Hinton et al.~\cite{hinton2015distilling} introduced knowledge distillation (KD) as a model compression technique in which a compact student network learns to mimic a larger teacher's output distribution. For generative models, sequence-level KD trains students on complete teacher outputs rather than only token-level distributions~\cite{kim2016sequence}. This paradigm has been widely adopted for language models: DistilBERT~\cite{sanh2019distilbert} compresses BERT while retaining 97\% of its performance, and MiniLLM~\cite{gu2024minillm} extends KD to autoregressive LLMs with a policy-level reverse KL objective. On-policy approaches such as GKD~\cite{agarwal2024gkd} further improve upon standard KD by training on the student's own generations and allowing alternative KL divergences. Closer to our setting, rationale- and explanation-trace distillation methods show that small models can benefit from LLM-generated reasoning supervision~\cite{hsieh2023distilling, mukherjee2023orca}. Preference-based alignment methods such as DPO~\cite{rafailov2023dpo} provide another route by optimizing pairwise teacher or human preferences rather than directly imitating a single teacher response. In low-resource RE, self-training and confidence calibration are also plausible alternatives; for example, PRiSM calibrates document-level RE logits with relation-aware scores when only a small amount of labeled data is available~\cite{choi2023prism}. In our work, we adopt a \emph{response-level} distillation strategy in which the teacher's evidence-grounded explanations and label outputs serve as silver training data for the student, combining structured reasoning transfer with the simplicity of standard SFT and avoiding the extra preference construction or confidence-threshold tuning required by these alternatives.

\paragraph{Parameter-efficient fine-tuning.}
LoRA~\cite{hu2022lora} and its quantized variant QLoRA~\cite{dettmers2023qlora} enable fine-tuning of billion-parameter models on consumer hardware by training only low-rank adapter matrices while keeping the base model weights frozen. We leverage QLoRA throughout our pipeline for both teacher and student training, making the entire workflow feasible on a single GPU.

\section{Approach}
\label{sec:approach}

\begin{figure}[t]
\centering
\includegraphics[width=1\linewidth]{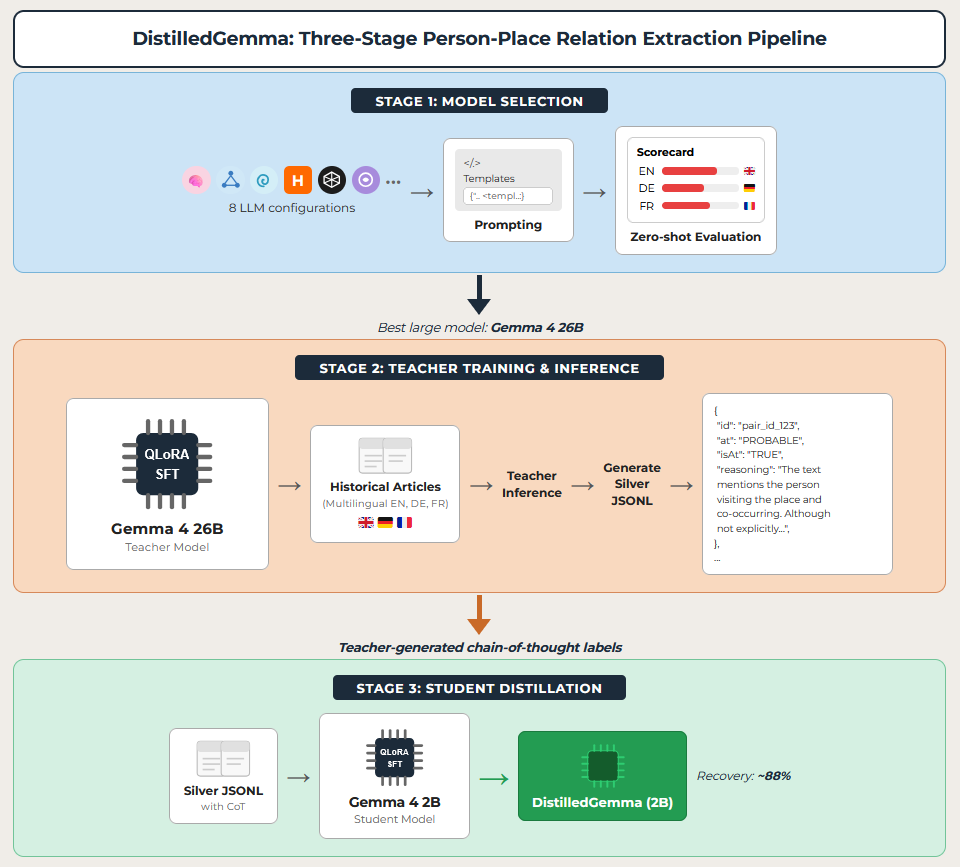}
\caption{Overview of the DistilledGemma three-stage pipeline. First, multiple LLM configurations and prompting strategies are evaluated to select the strongest teacher model. The teacher is then fine-tuned and used to generate chain-of-thought silver annotations, which are subsequently used to distill knowledge into a compact student model.}
\label{fig:pipeline}
\end{figure}

Our approach follows a three-stage pipeline, illustrated in Figure~\ref{fig:pipeline}: (1)~model selection through systematic experimentation, (2)~teacher fine-tuning and silver data generation, and (3)~student distillation.

\subsection{Stage 1: Model Selection and Prompt Engineering}
\label{sec:stage1}

We evaluated a diverse set of LLM families in multiple configurations:

\paragraph{Prompt engineering.} We designed five prompt variants ranging from a minimal zero-shot baseline to structured chain-of-thought templates that guide the model through explicit evidence assessment before emitting labels. The best-performing prompt (chain-of-thought), shown in Listing~\ref{lst:llm_prompt}, was used as the inference prompt for all LLM candidates to predict the \textit{at} and \textit{isAt} relations during base-model selection. It frames the model as a historian, defines the two relation labels, emphasizes OCR robustness and conservative inference, enforces the logical dependency between labels, and requires concise evidence-grounded JSON outputs.

\begin{lstlisting}[style=promptstyle,caption={Prompt template used for LLM inference, teacher generation, and student fine-tuning.},label={lst:llm_prompt}]
You are a historian working on person-location relations in multilingual historical
European newspapers.

Read carefully, expect OCR noise, and base every judgment only on the document text and
metadata given here.

Task:
- Classify relation "at" for the person and place. Allowed labels: TRUE, PROBABLE, FALSE
- Classify relation "isAt" for the person and place. Allowed labels: TRUE, FALSE

Relation meanings:
- "at" = the article gives textual evidence that the person was at the location at some
  point before publication.
- "isAt" = the article supports that the person was at the location in the article's
  immediate temporal horizon, meaning current, ongoing, or very recent relative to the
  publication date.

Decision guidance:
- Use TRUE for "at" only when the text gives clear evidence of presence.
- Use PROBABLE for "at" only when the text gives indirect, partial, or weak evidence that
  still points toward presence.
- Use FALSE for "at" when the article does not support presence.
- Use TRUE for "isAt" only when the article clearly places the person there in a current,
  ongoing, or very recent frame.
- Never use PROBABLE for "isAt". It must be TRUE or FALSE.
- If "at" is FALSE, then "isAt" must also be FALSE.
- If "isAt" is TRUE, then "at" must also be TRUE.

Rules:
- Do not use external knowledge.
- Do not infer more than the text warrants.
- Prefer FALSE when evidence is missing or too uncertain.
- Return JSON only.
- Use the exact person and place strings given for the current pair.
- Be robust to OCR noise and line-break artifacts in the text and in the entity mentions.
- Treat clearly equivalent mention surfaces as the same entity even if they differ by escaped line breaks, hyphenation, or minor OCR spelling noise.
- First write `at_explanation`, then decide `at`.
- Then write `isAt_explanation`, then decide `isAt`.
- Keep each explanation concise, at most 100 words.
- Each explanation must mention only evidence from the article, not your reasoning process.

Document language: {language}
Publication date: {publication_date}

Article text:
{article_text}

{pair_context}
\end{lstlisting}

\paragraph{Model families.} We evaluated models from four families: \textbf{Gemma}~\cite{gemma2team2024} (E2B, E4B, and 26B A4B), \textbf{Qwen}~\cite{qwen3} (2B to 9B), \textbf{Mistral}~\cite{mistral2024ministral} (3B), and \textbf{HY-MT}~\cite{zheng2025hymt} (1.8B). Models were served via vLLM~\cite{kwon2023vllm} for efficient batch inference.

\paragraph{Training strategies.} Beyond zero-shot prompting, we explored supervised fine-tuning (SFT) using LoRA~\cite{hu2022lora} on the HIPE training data. The SFT data was prepared by converting each labeled (person, place) pair into a chat-format record with the prompt as the user message and the gold JSON as the assistant response.

\subsection{Stage 2: Teacher Fine-Tuning}
\label{sec:stage2}

We evaluated \textbf{Gemma~4 26B A4B}~\cite{gemma4modelcard2026} using the same prompt as in the selection stage. We selected it as the teacher model because it achieved the strongest zero-shot macro-averaged recall (0.7255 across languages) and belongs to the same model family as the target student.

The teacher was fine-tuned using QLoRA~\cite{dettmers2023qlora} with the following configuration:
\begin{itemize}
    \item \textbf{LoRA}: rank $r{=}32$, $\alpha{=}64$, dropout $0.1$, applied to all attention and MLP projections (q/k/v/o/gate/up/down).
    \item \textbf{Training}: 3 epochs, batch size 1 with gradient accumulation 32 (effective batch size 32), cosine learning rate schedule with $\eta{=}2{\times}10^{-4}$ and 5\% warmup.
    \item \textbf{Data}: All labeled pairs from the newspaper and sandbox train splits across EN/DE/FR, with a 90/10 stratified train/dev split.
\end{itemize}

After training, the LoRA adapter was merged into the base model for efficient inference. The merged teacher was then used to generate silver-standard labels over the entire training corpus, producing chain-of-thought explanations alongside \textit{at} and \textit{isAt} predictions for each pair.

\subsection{Stage 3: Student Distillation}
\label{sec:stage3}

The student model (\textbf{Gemma~4 E2B}) was trained on the teacher-generated silver data using QLoRA with a lighter configuration:
\begin{itemize}
    \item \textbf{LoRA}: rank $r{=}16$, $\alpha{=}32$, dropout $0.05$.
    \item \textbf{Training}: 6 epochs (longer training compensates for the smaller model capacity), effective batch size 16.
    \item \textbf{Data}: Teacher-generated distilled JSONL, where the assistant messages contain the teacher's chain-of-thought with its predicted outputs rather than gold labels.
\end{itemize}

This \emph{response-level distillation} approach transfers not only the teacher's label decisions but also its reasoning patterns, enabling the student to learn structured inference over noisy historical text. The student learns to emulate the teacher's chain-of-thought process, which we hypothesize improves generalization compared to training on gold labels alone, as the latter lack detailed explanatory reasoning for many relation pairs.

For parameter accounting, we count only the deployable student model: the 26B teacher is used offline to create silver data, and the LoRA matrices are training-time adapters that are merged into the 2.3B base model for inference. Consequently, the efficiency profiles and compression ratios report the student as a 2.3B-parameter deployed model rather than summing teacher and student parameters or treating the adapters as a separate multi-billion-parameter component.

\subsection{Post-Processing}
\label{sec:postprocessing}

To address the imbalance between positive and negative relation pairs and ensure semantic consistency in model outputs, we applied rule-based post-processing to all generated predictions. Specifically, this step enforced: (1)~entity-type compatibility between the predicted relation and the entity types involved, (2)~relation-direction constraints ensuring the correct mapping of person-to-place, (3)~contextual trigger patterns that leverage surface-level cues in the article, and (4)~the logical constraint that $\text{isAt}{=}\textsc{TRUE} \Rightarrow \text{at}{=}\textsc{TRUE}$.

\section{Dataset}
\label{sec:data_stats}

The HIPE-2026 training corpus contains multilingual historical newspaper articles in German, English, and French with varying document lengths and entity distributions. Table~\ref{tab:dataset_overview} summarizes corpus-level statistics, while Tables~\ref{tab:relation_stats} and~\ref{tab:entity_stats} report relation and entity characteristics.

\begin{table}[h]
\centering
\small
\caption{Dataset overview by language.}
\label{tab:dataset_overview}
\begin{tabular}{lcccc}
\toprule
Language & Documents & Avg. Words & Avg. Characters\\
\midrule
German & 34 & 744.1 & 5000.9\\
English & 35 & 325.8 & 1802.7\\
French & 35 & 585.9 & 3635.0\\
\midrule
Total & 104 & -- & --\\
\bottomrule
\end{tabular}
\end{table}

\begin{table}[h]
\centering
\small
\caption{Relation label distribution across languages with overall statistics and class imbalance percentages. The label distribution exhibits significant class imbalance: for \textit{at}, FALSE accounts for $\sim$55\%, PROBABLE for $\sim$10\%, and TRUE for $\sim$35\%; for \textit{isAt}, FALSE dominates at $\sim$78\% versus TRUE at $\sim$22\%.}
\label{tab:relation_stats}
\begin{tabular}{lcccc}
\toprule
Label & German & English & French & Overall (percentage) \\
\midrule
\multicolumn{5}{c}{\textbf{at relation}} \\
\midrule
TRUE      & 135 & 125 & 181 & 441 (35.25\%) \\
FALSE     & 269 & 152 & 269 & 690 (55.16\%) \\
PROBABLE  & 62  & 30  & 28  & 120 (9.59\%) \\
\midrule
\multicolumn{5}{c}{\textbf{isAt relation}} \\
\midrule
TRUE      & 89  & 83  & 107 & 279 (22.30\%) \\
FALSE     & 377 & 224 & 371 & 972 (77.70\%) \\
\bottomrule
\end{tabular}
\end{table}

\begin{table}[h]
\centering
\small
\caption{Unique entity statistics.}
\label{tab:entity_stats}
\begin{tabular}{lcc}
\toprule
Language & Unique Persons & Unique Locations\\
\midrule
German & 174 & 217\\
English & 120 & 127\\
French & 209 & 237\\
\bottomrule
\end{tabular}
\end{table}

\section{Experiments}
\label{sec:experiments}

\subsection{Experimental Setup}
\label{sec:setup}

All models were served using vLLM~\cite{kwon2023vllm} with temperature $0.0$ and seed $42$ for output determinism.
We report macro- and micro-averaged recall following standard classification evaluation practice~\cite{sokolova2009systematic}. The logical constraint ($\text{isAt}{=}\textsc{TRUE} \Rightarrow \text{at}{=}\textsc{TRUE}$) was enforced as a post-processing step for all configurations.

\subsection{Results}
\label{sec:results}

Table~\ref{tab:results} presents the results across all eight model configurations.

\begin{table}[h]
\centering

\begin{tabular}{lccccc}
\toprule
\textbf{Model} & \textbf{EN} & \textbf{DE} & \textbf{FR} & \textbf{Macro Avg} & \textbf{Micro Avg} \\
\midrule

\textbf{Gemma~4 26B A4B}          & \textbf{0.7516} & \textbf{0.7239} & \textbf{0.7009} & \textbf{0.7255} & \textbf{0.8417} \\
Qwen3-4B             & 0.6452 & 0.6443 & 0.6493 & 0.6463 & 0.7626 \\
Gemma~4 E2B            & 0.6696 & 0.6336 & 0.6154 & 0.6395 & 0.7714 \\
Qwen3.5~9B           & 0.5641 & 0.6279 & 0.6268 & 0.6063 & 0.7912 \\
Mistral3~3B          & 0.5846 & 0.5852 & 0.5461 & 0.5720 & 0.6986 \\
Gemma~4 E4B            & 0.5458 & 0.5839 & 0.5830 & 0.5709 & 0.7804 \\
Qwen3.5~2B           & 0.5042 & 0.4886 & 0.4658 & 0.4862 & 0.5855 \\
HY-MT1.5~1.8B        & 0.4043 & 0.4219 & 0.3906 & 0.4056 & 0.3649 \\
\bottomrule
\end{tabular}

\small
\caption{Macro-averaged recall per language and overall metrics on the HIPE-2026 dataset. Models are sorted by macro-averaged recall. \textbf{Bold} indicates best in column.}
\label{tab:results}
\end{table}

Several key findings emerge from these results:

\paragraph{Large models dominate.} Gemma~4 26B A4B achieves the highest zero-shot macro-averaged recall (0.7255), confirming that model scale provides substantial advantages for this task. The model's strong multilingual pretraining contributes to consistent performance across all three languages.

\paragraph{Model scale is not the only factor.} Interestingly, smaller models with superior architecture can outperform larger but less capable ones. For instance, Qwen3-4B (macro 0.6463) and the base Gemma~4 E2B (macro 0.6395) both outperform the larger Qwen3.5~9B (macro 0.6063) and Gemma~4 E4B (macro 0.5709) in macro-averaged recall, suggesting that model architecture and pretraining data composition play a significant role alongside parameter count.

\paragraph{Small models are viable with distillation.} As shown in Section~\ref{sec:distill_analysis}, the distilled Gemma~4 E2B student achieves competitive performance that surpasses several larger zero-shot models, validating our distillation approach.

\section{Analysis}
\label{sec:analysis}

\subsection{Distillation Effectiveness}
\label{sec:distill_analysis}

Table~\ref{tab:distillation} quantifies the efficiency-accuracy trade-off achieved through distillation.

\begin{table}[h]
\centering

\begin{tabular}{lcccc}
\toprule
& \textbf{Deployed Params} & \textbf{Macro Avg} & Recovery Rate  & \textbf{Compression} \\
\midrule
Gemma~4 26B A4B (teacher) & 26B  & 0.7015 & ---  & --- \\
Gemma~4 E2B (distilled) & 2.3B & 0.6171 & $\sim$88\% & $\sim$11$\times$ \\
Gemma~4 E2B (zero-shot) & 2.3B & 0.5329  & $\sim$76\% & $\sim$11$\times$ \\
\bottomrule
\end{tabular}

\small
\caption{Comparison of teacher vs.\ distilled student performance on sandbox dev set. Recovery rate is the ratio of student to teacher macro-averaged recall. Parameter counts report deployed base-model size; LoRA adapters are merged for inference.}
\label{tab:distillation}

\end{table}

The distilled student achieves $\sim$88\% of the teacher's macro-averaged recall while being 11$\times$ smaller in deployed parameter count, confirming that teacher-generated reasoning patterns transfer effectively to the smaller model. These sandbox dev results are not directly comparable to the full test metrics reported in Table~\ref{tab:results}.

\subsection{Official Evaluation Results}
\label{sec:official}

In the official HIPE-2026 evaluation, our system (team12, ``whereami'') achieved competitive rankings across multiple profiles:

\begin{itemize}
    \item \textbf{Standard accuracy profile}: Ranked \textbf{3rd} among all teams with a mean profile score of \textbf{0.688}.
    \item \textbf{Binary accuracy profile}: Ranked \textbf{2nd} among all teams with a mean profile score of \textbf{0.8156}.
    \item \textbf{Balanced efficiency-accuracy profile}: Ranked \textbf{2nd} among all teams on both the standard and binary test sets, demonstrating an effective trade-off between model size and classification performance.
\end{itemize}

These results demonstrate that DistilledGemma provides a strong balance of accuracy and efficiency in several profiles while using substantially fewer parameters.

\subsection{Cross-Lingual Analysis}
\label{sec:crosslingual}

Performance varies across languages, with English generally achieving the highest scores and French exhibiting the greatest variance. German performance is notably consistent across model sizes, possibly because German historical newspapers in the HIPE dataset tend to feature cleaner OCR output and more explicit geographic references.

The distilled student shows the largest improvement on German, suggesting that the teacher's German reasoning patterns are particularly amenable to distillation. French performance proves more challenging to distill, likely due to the more complex syntactic structures characteristic of French historical prose.

\subsection{Error Analysis}
\label{sec:errors}

We identify three dominant error categories:

\begin{enumerate}
    \item \textbf{PROBABLE confusion}: The three-way \textit{at} classification proves challenging, with models frequently conflating PROBABLE and TRUE. This distinction requires nuanced assessment of evidence strength that smaller models struggle to capture.
    \item \textbf{False negatives on isAt}: The extreme class imbalance (77.7\% FALSE) biases all models toward negative predictions, particularly degrading recall on the minority TRUE class.
    \item \textbf{OCR-induced entity confusion}: In documents with severe OCR degradation, entity mentions are corrupted, leading to misidentification of both persons and places, which cascades into downstream relation classification errors.
\end{enumerate}

\subsection{Efficiency Considerations}
\label{sec:efficiency}

The Gemma~4 E2B student model can be served on a single consumer GPU with ${<}$8~GB VRAM, processing approximately 50 pairs per second via vLLM. This deployment count excludes the 26B teacher, which is used only offline for silver-data generation. In contrast, the 26B teacher requires a high-memory GPU. For large-scale historical document processing involving millions of articles, the 2.3B student provides a practical solution with acceptable performance degradation, while the 26B teacher remains preferable when accuracy is paramount and computational resources are available.

\section{Conclusion}
\label{sec:conclusion}

We presented DistilledGemma, a knowledge distillation pipeline for person-place relation extraction from multilingual historical newspaper texts. Through systematic evaluation of eight LLM configurations, we demonstrated that supervised fine-tuning of large models yields the strongest performance, and that response-level distillation effectively transfers chain-of-thought reasoning patterns to compact models.

Our key findings are: (1)~QLoRA fine-tuning enables the selected teacher to generate task-specific silver data, and response-level distillation transfers much of that performance to the student; (2)~knowledge distillation from a 26B teacher to a 2.3B student recovers ${\sim}$88\% of teacher performance at $\nicefrac{1}{11}$ the parameter cost; and (3)~chain-of-thought distillation, where the student learns the teacher's reasoning process rather than labels alone, combined with rule-based post-processing, provides an effective pathway to competitive performance with limited computational resources.

In the official HIPE-2026 evaluation, DistilledGemma ranked 3rd in the standard accuracy profile and 2nd in both the binary accuracy and balanced efficiency-accuracy profiles, demonstrating that strong performance on multilingual historical relation extraction can be achieved without excessive computational expenditure.

Future work includes exploring on-policy self-distillation methods~\cite{agarwal2024gkd} that could further narrow the teacher-student performance gap, investigating encoder-based discriminative models as an alternative lightweight approach, and extending the framework to additional historical document domains.

\small
\bibliography{references}

\end{document}